\documentclass[sigconf]{acmart}

\usepackage{booktabs}
\usepackage{color}
\usepackage{pifont}
\usepackage{multirow}
\usepackage{subfig}

\AtBeginDocument{%
  }

\copyrightyear{2024}
\acmYear{2024}
\setcopyright{rightsretained}
\acmConference[ICMR '24]{Proceedings of the 2024 International Conference on Multimedia Retrieval}{June 10--14, 2024}{Phuket, Thailand}
\acmBooktitle{Proceedings of the 2024 International Conference on Multimedia Retrieval (ICMR '24), June 10--14, 2024, Phuket, Thailand}\acmDOI{10.1145/3652583.3658044}
\acmISBN{979-8-4007-0619-6/24/06}

\makeatletter
\gdef\@copyrightpermission{
  \begin{minipage}{0.3\columnwidth}
   \href{https://creativecommons.org/licenses/by/4.0/}{\includegraphics[width=0.90\textwidth]{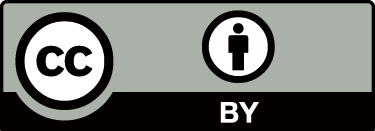}}
  \end{minipage}\hfill
  \begin{minipage}{0.7\columnwidth}
   \href{https://creativecommons.org/licenses/by/4.0/}{This work is licensed under a Creative Commons Attribution International 4.0 License.}
  \end{minipage}
  \vspace{5pt}
}
\makeatother

\begin{document}

\title{Multimodal Prototype-Enhanced Network for Few-Shot Action Recognition}

\author{Xinzhe Ni}
\email{nxz22@mails.tsinghua.edu.cn}
\affiliation{%
  \institution{Tsinghua University}
  \city{Shenzhen}
  \country{China}
}

\author{Yong Liu}
\email{liu-yong20@mails.tsinghua.edu.cn}
\affiliation{%
  \institution{Tsinghua University}
  \city{Shenzhen}
  \country{China}
}

\author{Hao Wen}
\email{wenh22@mails.tsinghua.edu.cn}
\affiliation{%
  \institution{Tsinghua University}
  \city{Shenzhen}
  \country{China}
}

\author{Yatai Ji}
\email{jyt21@mails.tsinghua.edu.cn}
\affiliation{%
  \institution{Tsinghua University}
  \city{Shenzhen}
  \country{China}
}

\author{Jing Xiao}
\email{xiaojing661@pingan.com.cn}
\affiliation{%
  \institution{Ping An Insurance (Group) Company of China}
  \city{Shenzhen}
  \country{China}
}

\author{Yujiu Yang}
\authornote{Corresponding author.}
\email{yang.yujiu@sz.tsinghua.edu.cn}
\affiliation{%
  \institution{Tsinghua University}
  \city{Shenzhen}
  \country{China}
}

\renewcommand{\shortauthors}{Xinzhe Ni et al.}
\begin{abstract}
  Current methods for few-shot action recognition mainly fall into the metric learning framework following ProtoNet, which demonstrates the importance of prototypes. Although they achieve relatively good performance, the effect of multimodal information is ignored, e.g. label texts. In this work, we propose a novel \textbf{M}ultim\textbf{O}dal P\textbf{R}ototype-E\textbf{N}hanced Network (MORN), which uses the semantic information of label texts as multimodal information to enhance prototypes. A CLIP visual encoder and a frozen CLIP text encoder are introduced to obtain features with good multimodal initialization. Then in the visual flow, visual prototypes are computed by a visual prototype-computed module. In the text flow, a semantic-enhanced (SE) module and an inflating operation are used to obtain text prototypes. The final multimodal prototypes are then computed by a multimodal prototype-enhanced (MPE) module. Besides, we define a \textbf{PR}ototype S\textbf{I}milarity \textbf{D}iff\textbf{E}rence (PRIDE) to evaluate the quality of prototypes, which is used to verify our improvement on the prototype level and effectiveness of MORN. We conduct extensive experiments on four popular few-shot action recognition datasets: HMDB51, UCF101, Kinetics and SSv2, and MORN achieves state-of-the-art results. When plugging PRIDE into the training stage, the performance can be further improved.
\end{abstract}

\ccsdesc[500]{Computing methodologies~Activity recognition and understanding}


\keywords{Few-shot action recognition, Prototype, Multimodal understanding}


\maketitle

\begin{figure}[t]
  \centering
  \subfloat[Prototype-based strategy.]{
    \includegraphics[width=1\linewidth]{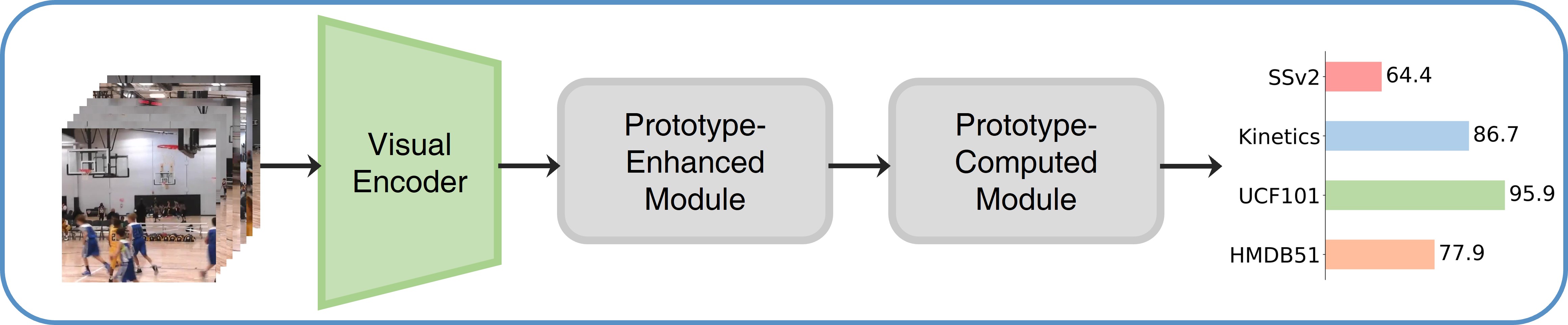}
    \label{fig1-a}
    }
    
  \subfloat[Multimodal-based strategy.]{
    \includegraphics[width=1\linewidth]{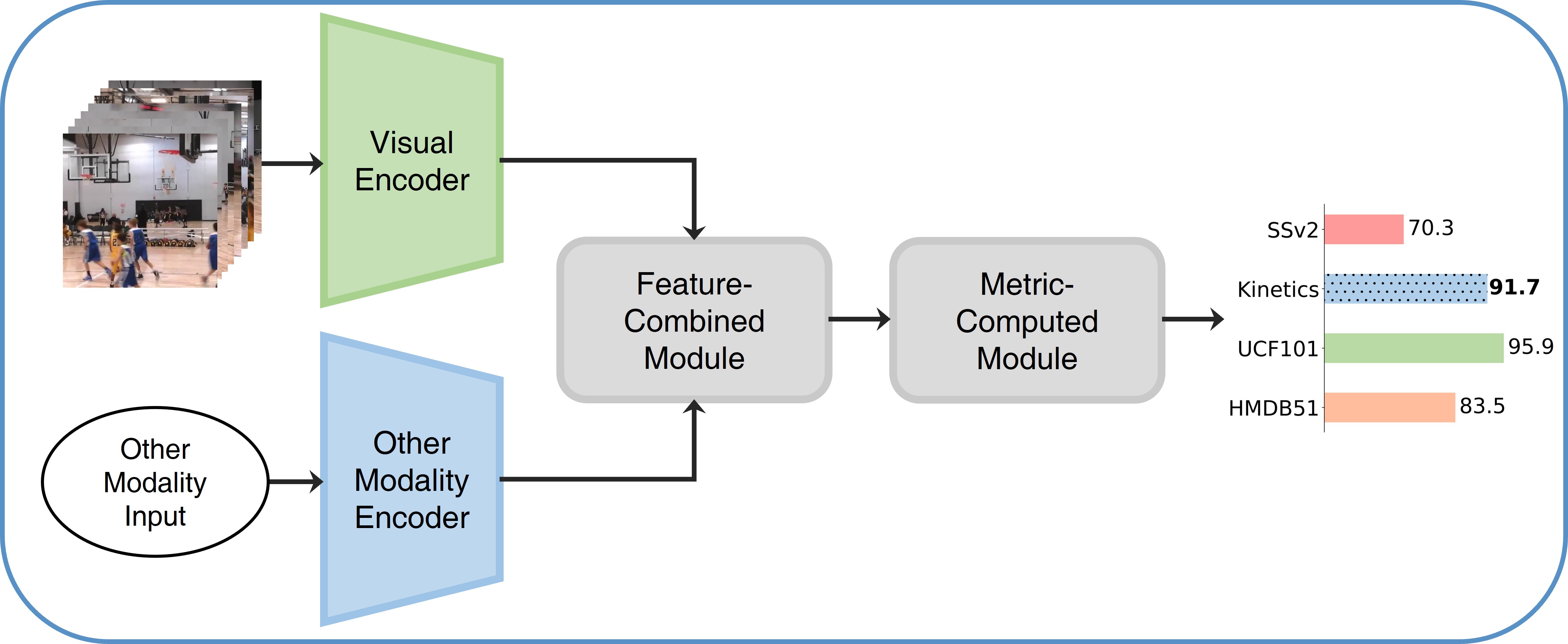}
    \label{fig1-b}
    }
    
  \subfloat[Our multimodal prototype-enhanced strategy.]{
    \includegraphics[width=1\linewidth]{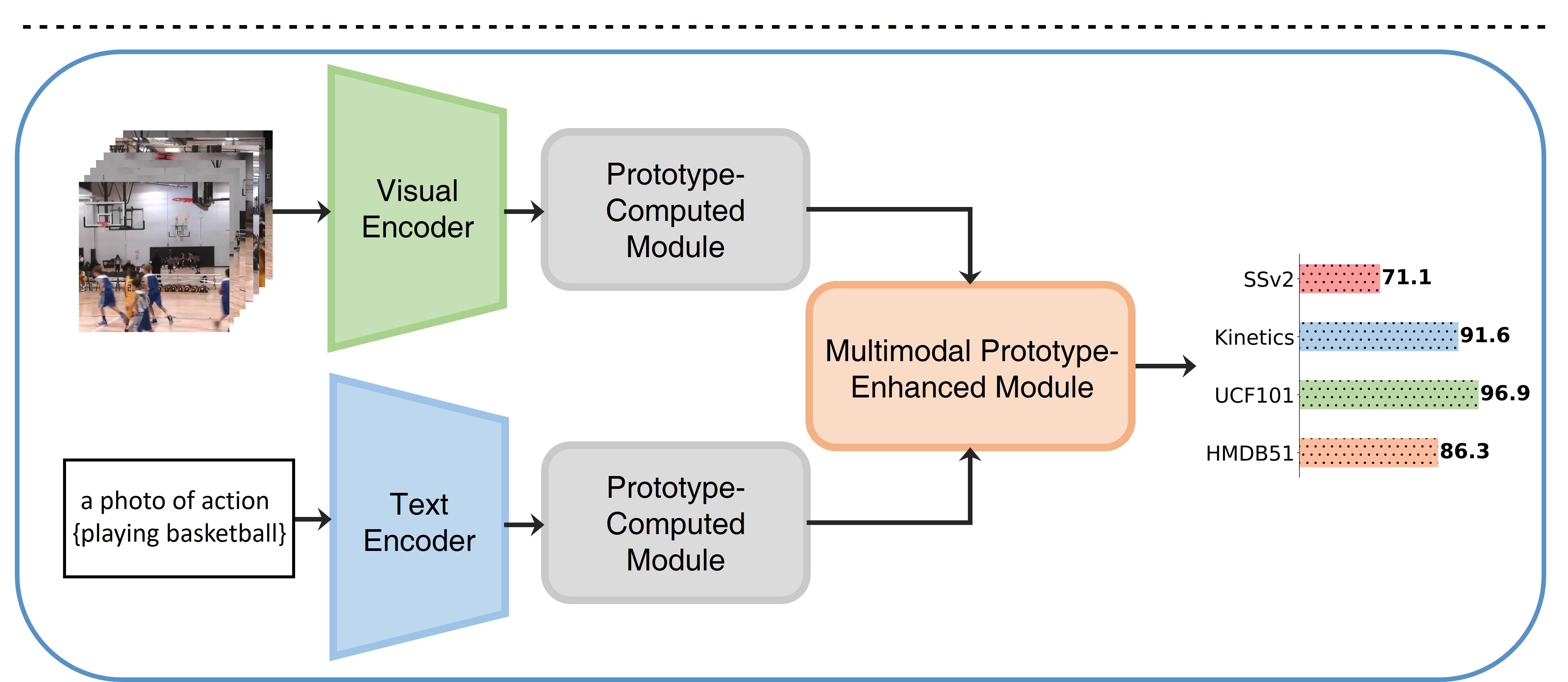}
    \label{fig1-c}
    }
  \caption{Existing metric-learning strategy (a) and (b) and our multimodal prototype-enhanced strategy (c) for few-shot action recognition. Results on TRX~\cite{perrett2021temporal} with CLIP visual encoder, multimodal feature-enhanced strategy and multimodal prototype-enhanced strategy are shown.}
  \label{fig1}
\end{figure}

\section{Introduction}
\label{sec:intro}
Few-shot action recognition is a challenging problem in computer vision because of the scarcity of labeled samples and the complicated temporal evolution of videos. Most of the methods~\cite{bishay2019tarn, kumar2019protogan, zhu2018compound, zhang2020few-, fu2020depth, cao2020few, zhang2020few, zhang2021learning, perrett2021temporal, thatipelli2022spatio, wang2022hybrid, huang2022compound, liu2021staf, zhu2021few, wu2022motion} for few-shot action recognition mainly fall into the metric-learning framework, whose key idea is to learn a proper metric space tuned to a particular task. Among various metric-learning works, ProtoNet~\cite{snell2017prototypical} has been followed by many works and has far-reaching influence. It proposes Prototype as a representation of each category in a few-shot learning scenario. For the scarcity of samples, it is likely that the overfitting problem will arise with a complicated matching strategy. With prototypes, however, the model reflects a simpler inductive bias and works better in a limited-data regime, \emph{i.e.} few-shot learning scenario. Since we notice the importance of prototypes, it is natural to ask: \emph{how to obtain representative prototypes}?

Intuitively, it is useful to directly introduce a certain prototype-enhanced module as shown in Figure \ref{fig1-a}. Methods with prototype-based strategy~\cite{kumar2019protogan, perrett2021temporal, zhu2021few} expand the concept of prototype, which is not a simple average of video features. However, the information provided by a single modality is limited especially in the few-shot scenario, and they ignore the effect of multimodal information.

As we know, video is a multimodal carrier rather than a series of RGB images. Multimodal information generally yields better results since different modalities provide complementary information, and is proved by~\cite{huang2021makes}. Therefore, another strategy is to use multimodal information to obtain a better distance metric as shown in Figure \ref{fig1-b}. We select label texts as another modality inspired by ActionCLIP~\cite{wang2021actionclip}. There is abundant semantic information in label texts to assist the few-shot classification. For example, if one is given several short videos of ``lifting up one end of something'' without being told what action it is, he may not find the same pattern in these videos in a short time. Instead, it is much easier to distinguish them from other actions when knowing the exact label. Based on this fact, we preliminarily combine the semantic information of label texts on the feature level. Compared to unimodal prototype-enhanced strategy in Figure \ref{fig1-a}, the model with multimodal feature-enhanced strategy achieves further improvement, which verifies the effectiveness of multimodal information. However, methods with multimodal-based strategy~\cite{zhang2020few-, fu2020depth} ignore the importance of prototypes and only combine the two modalities on the feature level. As a result, there is still large room for improvement with a multimodal prototype-enhanced strategy. 

Considering both prototype and multimodal information, we propose a novel \textbf{M}ultim\textbf{O}dal P\textbf{R}ototype-E\textbf{N}hanced Network (MORN) based on our multimodal prototype-enhanced strategy as shown in Figure \ref{fig1-c}. It is simple but well-performed, and can be applied to methods producing prototypes. For example on TRX~\cite{perrett2021temporal}, we firstly obtain visual and text features by a CLIP~\cite{radford2021learning} visual encoder and a frozen CLIP text encoder. Then, visual prototypes are computed by TRX baseline, and text prototypes are computed by a semantic-enhanced (SE) module and an inflating operation. Finally, we combine visual prototypes and text prototypes to obtain representative prototypes in a multimodal prototype-enhanced (MPE) module. The final multimodal prototypes are used during the distance computation with query videos. Compared to multimodal feature-enhanced strategy in Figure \ref{fig1-b}, MORN achieves better or comparable improvement, which verifies the importance of multimodal prototype-enhanced strategy.

To verify that our MORN achieves improvement on prototypes rather than other aspects, there is a strong need to evaluate the quality of prototypes with our MORN. Inspired by~\cite{zhang2021prototype}, we define \textbf{PR}ototype S\textbf{I}milarity \textbf{D}iff\textbf{E}rence (PRIDE). Firstly, we compute the real prototype of each category of all samples in the meta-test stage, and the similarity of current prototypes in an episode with all real prototypes. Then, the difference of similarity between its own category and the average of other categories is used to evaluate the quality of prototypes. With experiments on PRIDE, we are convinced that our MORN achieves improvement on the prototype level. Since PRIDE indicates the quality of prototypes, it can also be used to assist the training process. Therefore, we plug PRIDE into the training stage of MORN and achieve further performance gains as we had expected.

Our contributions can be summarized as follows: (1) We use the semantic information of label texts to enhance prototypes and propose a simple but well-performed Multimodal Prototype-Enhanced Network (MORN). (2) We define Prototype Similarity Difference (PRIDE) to verify improvement on the prototype level and effectiveness of MORN. When plugging PRIDE into the training stage, MORN can achieve further performance gains. (3) We conduct extensive experiments on four popular action recognition datasets and the application of PRIDE. Methods with MORN achieve state-of-the-art results on HMDB51, UCF101, Kinetics and SSv2.

\section{Related Work}
\label{sec:related work}

\subsection{Few-Shot Image Classification}

Few-shot image classification methods can be widely divided into three categories: augmentation-based, optimization-based and metric-based methods. 

\noindent\textbf{Augmentation-based methods.} The objective of these methods is to use augmentation techniques or extra data to increase samples for training, and improve the diversity of data. Some prior attempts are intuitive, including~\cite{perez2017effectiveness, ratner2017learning}. Besides, some works focus on the semantic feature level, including~\cite{chen2018semantic, chen2019multi}. Rather than applying augmentation techniques, ~\cite{pahde2019self, pahde2021multimodal} uses a GAN~\cite{goodfellow2020generative} architecture to generate extra images based on the text description to compensate for the lack of data. 

\noindent\textbf{Optimization-based methods.} The objective of these methods is to train a model under the meta-learning framework so that it can adapt to novel tasks with only a few optimization steps. ~\cite{andrychowicz2016learning, gui2018few, lee2019meta, liu2020ensemble, ravi2016optimization} use the meta-learner as an optimizer. MAML~\cite{finn2017model} and its variants~\cite{antoniou2018train, jamal2019task, sun2019meta} aim to learn a robust model initialization for the adaptation.

\noindent\textbf{Metric-based methods.} The objective of these methods is to learn feature embeddings under a certain distance metric with a better generalization ability. Samples of novel categories can be accurately classified via a nearest neighbor classifier with different distance metrics such as cosine similarity~\cite{vinyals2016matching, ye2020few}, Euclidean distance~\cite{snell2017prototypical, yoon2019tapnet}, non-linear metric with CNNs~\cite{sung2018learning, hou2019cross, li2019finding} or graph neural networks~\cite{kim2019edge, garcia2017few, yang2020dpgn}. Our work falls into the metric-learning framework and aims to solve a more complicated few-shot video action recognition problem.

\begin{figure*}
  \centering
  \includegraphics[width=0.85\linewidth]{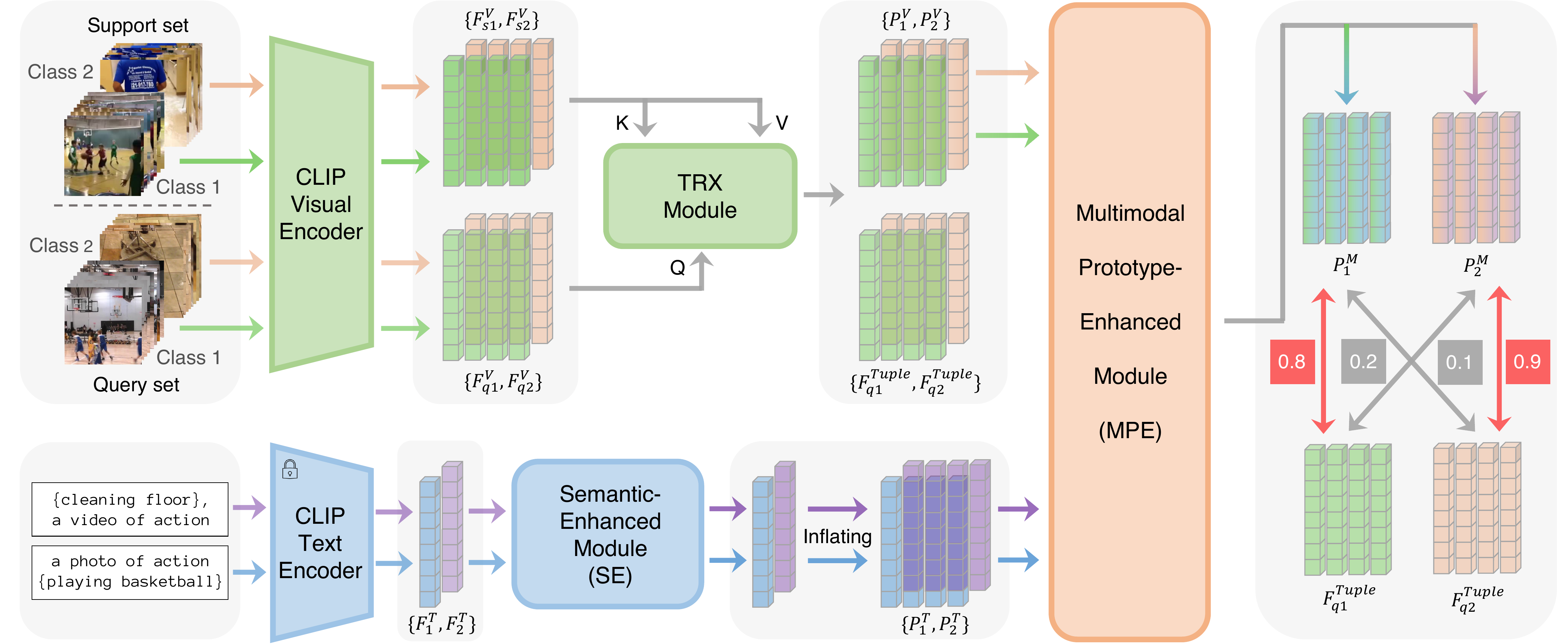}
  \caption{
 Overview of our proposed MORN on TRX on a 2-way 1-shot problem with 1 video for each category in the query set. In the visual flow, a CLIP visual encoder is first introduced on videos with $L$ frames to obtain video features. Then, support video features are passed to the Temporal-Relational CrossTransformer (TRX) module to compute visual prototypes. In the text flow, a frozen CLIP text encoder is first introduced on the prompted label texts. Then, the semantic features of label texts are passed to a semantic-enhanced (SE) module and are inflated as text prototypes. The visual prototypes and the text prototypes are combined through a multimodal prototype-enhanced (MPE) module.} 
  \label{fig2}
\end{figure*}

\subsection{Few-Shot Action Recognition}

Most of the methods for few-shot action recognition mainly fall into the metric-learning framework. More specifically, we divide them into three categories: feature-based, multimodal-based and prototype-based methods. 

\noindent\textbf{Feature-based methods.} The objective of these methods is to obtain reliable video features or apply an effective alignment strategy. CMN~\cite{zhu2018compound}, TARN~\cite{bishay2019tarn}, OTAM~\cite{cao2020few} and ARN~\cite{zhang2020few} make preliminary attempts to obtain reliable video features for classification. ITANet~\cite{zhang2021learning} introduces a video representation based on a self-attention mechanism and an implicit temporal alignment. MASTAF~\cite{liu2021staf} uses self-attention and cross-attention mechanisms to obtain reliable spatial-temporal features. MT-FAN~\cite{wu2022motion} designs a motion modulator and a segment attention mechanism, and conducts a temporal fragment alignment. STRM~\cite{thatipelli2022spatio} proposes a local patch-level and a global frame-level enrichment module based on TRX. Similarly, HyRSM~\cite{wang2022hybrid} proposes a hybrid relation module to enrich video features. To learn temporal and spatial associations, HCL~\cite{zheng2022few} applies a mixed-supervised hierarchical contrastive learning strategy. MoLo~\cite{wang2023molo} applies a long-short contrastive objective to connect local frame features to the global context, and uses a motion autodecoder for motion information extraction.

\noindent\textbf{Prototype-based methods.} Prototype is a representation of each category, and is proposed by ProtoNet~\cite{snell2017prototypical}. The objective of these methods is to compute representative prototypes by some prototype-enhanced strategies with a single modality. ProtoGAN~\cite{kumar2019protogan} generates extra samples by using a conditional GAN with category prototypes. PAL~\cite{zhu2021few} matches a prototype with all the query samples instead of matching a query sample with the space of prototypes. TRX~\cite{perrett2021temporal} applies CrossTransformer~\cite{doersch2020crosstransformers} to a few-shot action recognition scenario to obtain query-specific category prototypes. Based on HyRSM~\cite{wang2022hybrid}, GgHM~\cite{xing2023boosting} pays attention to the prototype construction and matching, thus outperforming HyRSM baseline. The improvement of GgHM indicates the importance of prototypes preliminary.

\noindent\textbf{Multimodal-based methods.} The objective of these methods is to use multimodal information to assist classification. AMeFu-Net~\cite{fu2020depth} introduces depth information as another modality to assist classification. CMMN~\cite{zhang2020few-} also uses the semantic information of label texts as ours and is similar to our multimodal setting, while it fails to apply a feature extractor with a good multimodal initialization. CPMT~\cite{huang2022compound} uses object features as multimodal information and proposes a compound matching strategy, and achieves good performance. Recently, some works have paid attention to the combination of multimodal information and prototypes: CLIP-FSAR~\cite{wang2023clip} uses a video-text contrastive objective and MA-CLIP~\cite{xing2023multimodal} adds lightweight adapters to adapt the CLIP model to the few-shot action recognition task. Our work also uses multimodal information and computes representative prototypes. Furthermore, we pay more attention on the remarkable effect of the prototype-enhanced strategy and defines Prototype Similarity Difference (PRIDE) to verify the improvement on the prototype level.

\section{Method}
\label{sec:method}

\subsection{Problem formulation}
\label{problem formualtion}

There are two sets in the $N$-way $K$-shot few-shot scenario: a support set $S^{N}_{K}=\{s^{1}_{1},\cdots,s^{1}_{K},\cdots,s^{N}_{1},\cdots,s^{N}_{K}\}$ and a query set $Q^{N}_{M}=\{q^{1}_{1},\cdots,q^{1}_{M},\cdots,q^{N}_{1},\cdots,q^{N}_{M}\}$, where $N$, $K$ and $M$ denote the number of action categories, the number of videos each category in the support set, and the number of videos each category in the query set respectively. The objective of few-shot action recognition is to learn a strong model with only a few video samples. Specifically, the model classifies a completely novel query action to the right category by matching it with the most similar video in the support set. Meanwhile, the whole dataset can be also divided into a base dataset $C_{base}=\{(x_{i},y_{i})\}^{|C_{base}|}_{i=1}$ and a novel dataset $C_{novel}=\{(x_{i},y_{i})\}^{|C_{novel}|}_{i=1}$, where $y_{i}$ is the action category of a video sample $x_{i}$. $C_{base}$ and $C_{novel}$ contain samples in the meta-training stage and meta-test stage respectively. Note that the categories of $C_{base}$ and $C_{novel}$ are label-wise non-overlapping, \emph{i.e. $C_{base}\,\cap\,C_{novel}=\emptyset$}. To make full use of the few samples, we follow the episode training manner~\cite{vinyals2016matching}. Each episode contains $K$ videos in the support set and $M$ videos in the query set of $N$ categories.

\subsection{MORN}

Considering both prototype and multimodal information, we propose MORN. The overall architecture is shown in Figure \ref{fig2}, including a visual flow, a text flow and a multimodal prototype-enhanced (MPE) module.

\noindent\textbf{Visual flow.} For each input video, we uniformly sample $L$ frames as in~\cite{wang2016temporal}. The video $s^{n}_{k}$ in the support set and $q^{n}_{m}$ in the query set can be further denoted as:
\begin{equation}
\begin{aligned}
\label{eq:eq4}
s^{n}_{k}&=\{s^{n}_{k1},\cdots,s^{n}_{kL}\},\\q^{n}_{m}&=\{q^{n}_{m1},\cdots,q^{n}_{mL}\},
\end{aligned}
\end{equation}
where $s^{n}_{k}$, $q^{n}_{m} \in \mathbb{R}^{L \times H \times W \times 3}$. $H$ and $W$ denote the height and width of an image. Then, we apply a pre-trained CLIP visual encoder for better multimodal initialization. The visual features of each frame $s^{n}_{ki} \in \mathbb{R}^{H \times W \times 3}$ in the support set and $q^{n}_{mi} \in \mathbb{R}^{H \times W \times 3}$ in the query set are defined as:
\begin{equation}
\begin{aligned}
\label{eq:eq5}
F^{V}_{sk}&=\{f_{v}(s^{n}_{k1}),\cdots,f_{v}(s^{n}_{kL})\},\\F^{V}_{qm}&=\{f_{v}(q^{n}_{m1}),\cdots,f_{v}(q^{n}_{mL})\},
\end{aligned}
\end{equation}
where $F^{V}_{sk}$, $F^{V}_{qm} \in \mathbb{R}^{L \times d}$. The visual features of videos in the support set and the query set are denoted as:
\begin{equation}
\begin{aligned}
\label{eq:eq6}
F^{V}_{s}&=\{F^{V}_{s1},\cdots,F^{V}_{s(NK)}\},\\F^{V}_{q}&=\{F^{V}_{q1},\cdots,F^{V}_{q(NM)}\},
\end{aligned}
\end{equation}
where $F^{V}_{s} \in \mathbb{R}^{NK \times L \times d}$ and $F^{V}_{q} \in \mathbb{R}^{NM \times L \times d}$. Then, we sample tuples of video frames $F^{Tuple}_{q}=\{F^{Tuple}_{q1},\cdots,F^{Tuple}_{q(NM)}\}$ and compute visual prototypes of each episode through TRX:
\begin{equation}
\label{eq:eq7}
P^{V}=TRX(F^{V}_{q},F^{V}_{s},F^{V}_{s}),
\end{equation}
where $F^{Tuple}_{q}, P^{V} \in \mathbb{R}^{NM \times C^{\omega}_{L} \times d_{p}}$, and $C^{\omega}_{L}$ is the combinatorial number.

\noindent\textbf{Text flow.} In the training stage for each sample $(x_{i}, y_{i}) \in C_{base}$, we firstly make $n_{temp}$ discrete templates $P_{temp}=\{p^{temp}_{1},\cdots,p^{temp}_{n_{temp}}\}$ of $y_{i}$ as text prompts. Then, we apply a CLIP text tokenizer to obtain tokenized text sequences $T_{i}$:
\begin{equation}
\begin{aligned}
\label{eq:eq8}
T_{i}&=Tokenizer(p^{temp}_{j}||y_{i}),\\j&=1,\cdots,n_{temp},\,i=1,\cdots,N,
\end{aligned}
\end{equation}
where $y_{i}$ denotes the action category of a video sample $x_{i}$ as described in Section \ref{problem formualtion}, and $||$ denotes a concatenation operation. In the practical meta-training and meta-test stage, the template is randomly selected once per episode. Then, the features of prompted label texts are obtained by a frozen CLIP text encoder:
\begin{equation}
\label{eq:eq9}
F^{T}_{i}=f_{t}(T_{i}),\,\{ i\,|\,x_{i} \in S^{N}_{K}\,\cup\,Q^{N}_{M}\},
\end{equation}
where $F^{T}_{i} \in \mathbb{R}^{1 \times d}$. To obtain text features with more reliable semantic information, we further apply a semantic-enhanced (SE) module $g(\cdot)$:
\begin{equation}
\label{eq:eq10}
P^{T}_{i}=g(F^{T}_{i}),
\end{equation}
where $P^{T}_{i} \in \mathbb{R}^{1 \times d_{p}}$ and a multi-head attention mechanism with 4 heads is used as $g(\cdot)$. So we obtain $\{P^{T}_{1},\cdots,P^{T}_{NM}\}$ in each episode. Since different frames of videos in the same category have the same label, we simply inflate $P^{T}_{i}$ of the same category to keep the same dimension with visual prototypes. Then, we obtain text prototypes $P^{T}$:
\begin{equation}
\label{eq:eq11}
P^{T}=\{P^{T}_{1},\cdots,P^{T}_{1},\cdots,P^{T}_{N},\cdots,P^{T}_{N}\},
\end{equation}
where $P^{T} \in \mathbb{R}^{NM \times C^{\omega}_{L} \times d_{p}}$.

\noindent\textbf{Multimodal prototype-enhanced module.}
To use multimodal information to enhance prototypes, we propose a simple but well-performed multimodal prototype-enhanced (MPE) module. The choice of the MPE module is flexible containing weighted average, multi-head attention and so on. Here, we apply weighted average as our default setting:
\begin{equation}
\label{eq:eq12}
P^{M}=(1-\lambda) P^{V}+\lambda P^{T},
\end{equation}
where $P^{M} \in \mathbb{R}^{NM \times C^{\omega}_{L} \times d_{p}}$ and $\lambda$ is the multimodal enhanced hyper-parameter. The multimodal prototypes are used as the final prototypes. Then, distances are computed between multimodal prototypes and $F^{Tuple}_{q}$, and are passed as logits to a cross-entropy loss.

\subsection{PRIDE}
\label{PRIDE}
Denote the number of categories in the meta-test stage as $N_{novel}$. We first compute the real prototype in a 1-way scenario by averaging all prototypes of the same category $i$ in the novel dataset, which is based on~\cite{zhang2021prototype}:
\begin{equation}
\label{eq:eq13}
P^{real}_{i}=\frac{1}{|C^{i}_{novel}|} \sum_{(x,y) \in C^{i}_{novel}} P_{i}(x),
\end{equation}
where $P_{i}(x)$ is the query-specific category prototype of video $x$ and $C^{i}_{novel}$ is the i-th category in $C_{novel}$. The whole set of real prototypes is denoted as:
\begin{equation}
\label{eq:eq14}
P^{real}=\{P^{real}_{1},\cdots,P^{real}_{N_{novel}}\}.
\end{equation}

Then, the cosine similarity can be computed between a given $P_{i}(x)$ and $P^{real}_{j}$:
\begin{equation}
\label{eq:eq15}
Sim_{j}=sim(P_{i}(x),P^{real}_{j}),\,j=1,\cdots,N_{novel},
\end{equation}
where $sim(\cdot,\cdot)$ is the cosine similarity operation. The similarity of other categories is denoted as:
\begin{equation}
\label{eq:eq16}
Sim_{other}=\frac{1}{N_{novel}-1} \sum_{1\le j\neq i\le N_{novel}} Sim_{j}.
\end{equation}

We can now denote PRIDE as: 
\begin{equation}
\label{eq:eq17}
PRIDE_{i}(x) = Sim_{i}-Sim_{other},\, (x,y) \in C^{i}_{novel}.
\end{equation}

According to the computation process, PRIDE can accurately reflect the category separability of prototypes. Thus, PRIDE can be used to verify the improvement on the prototype level and the effectiveness of our MORN. A higher PRIDE value means more representative prototypes with a better discriminating ability.

\begin{table*}[ht]
  \centering
  \caption{State-of-the-art comparison on four few-shot action recognition datasets in terms of classification accuracy. ``Weighted average'' and ``Multi-head Attention'' denotes different choices of the MPE module. *$^{M}$ denotes methods with our MORN. The bold font and underline indicate the best and second-best results respectively.}
  \resizebox{\linewidth}{!}{
  \begin{tabular}{c|c|c|c|ccccc}
  \toprule
  Method & Reference & Strategy & Backbone & HMDB51 & UCF101 & Kinetics & SSv2 & Average \\\midrule
  ARN~\cite{zhang2020few}  & ECCV' 20 & \multirow{6}{*}{Feature-based} & C3D & 60.6 & 83.1 & 82.4 & - & 75.4    \\\cline{4-3}
  OTAM~\cite{cao2020few} & CVPR' 20 &  & \multirow{7}{*}{ResNet-50} & 68.0 & 88.9 & 85.8 & 52.3 & 73.8 \\
  HyRSM~\cite{wang2022hybrid} & CVPR' 22 &  &  & 76.0 & 94.7 & 86.1 & 69.0 & 81.5 \\
  MT-FAN~\cite{wu2022motion} & CVPR' 22 &  &  & 74.6 & 95.1 & 87.4 & 60.4 & 79.4 \\
  STRM~\cite{thatipelli2022spatio} & CVPR' 22 &  &  & 77.3 & 96.9 & 86.7 & 68.1 & 82.3 \\
  HCL~\cite{zheng2022few} & ECCV' 22 &  &  & 76.3 & 93.9 & 85.8 & 64.9 & 80.2 \\
  MoLo~\cite{wang2023molo}  & CVPR' 23 &  &  & 76.1 & 95.1 & 85.1 & 69.6 & 81.5 \\\midrule
  PAL~\cite{zhu2021few}  & BMVC' 21 & \multirow{4}{*}{Prototype-based} & \multirow{4}{*}{ResNet-50} & 75.8 & 95.2 & 87.1 & 62.6 & 80.2 \\
  TRX~\cite{perrett2021temporal}  & CVPR' 21 &  &  & 75.6 & 96.1 & 85.9 & 64.6 & 80.6 \\
  GgHM~\cite{xing2023boosting}  & ICCV' 23 &  &  & 76.9 & 96.3 & 87.4 & 69.2 & 82.5 \\\midrule
  CMMN~\cite{zhang2020few-} & PR' 20 & \multirow{3}{*}{Multimodal-based}  & C3D + MLP & 70.5   & 87.7   & -        & -    & 79.1    \\\cline{4-3}
  AMeFu-Net~\cite{fu2020depth}  & ACM MM' 20 &  & ResNet-50 + ResNet-50 & 75.5 & 95.5 & 86.8 & - & 85.9 \\\cline{4-3}
  CPMT~\cite{huang2022compound} & ECCV' 22 &  & ResNet-50 + Mask-RCNN & 85.1   & 92.3 & 87.9 & \underline{73.5} & 84.7 \\\midrule
  CLIP-FSAR~\cite{wang2023clip} & ArXiv' 23 & \multirow{6}{*}{\begin{tabular}{@{}c@{}} Prototype-based \& \\Multimodal-based \end{tabular}} & \multirow{6}{*}{\begin{tabular}{@{}c@{}} CLIP ResNet-50 + \\CLIP text encoder \end{tabular}} & 80.7   & \underline{97.0} & 92.0 & 62.8 & 83.1 \\
  MA-CLIP~\cite{xing2023multimodal} & ArXiv' 23 &  &  & 82.1   & 96.6 & 93.0 & 63.6 & 83.8 \\\cline{1-2}\cline{5-9}
  Ours: STRM$^M$ (Weighted average) & \multirow{4}{*}{-} &  &   & 84.4   & 94.6   & 90.5     & \textbf{73.6} & 85.8    \\ 
  Ours: TRX$^M$ (Weighted average) &  &  &  & \underline{86.3} & 96.9 & 91.6 & 71.1 & 86.5 \\
  Ours: STRM$^M$ (Multi-head Attention) &  &  &  & 86.2 & 96.4 & \underline{93.8} & 71.9 & \underline{87.1} \\
  Ours: TRX$^M$ (Multi-head Attention) &  &  &  & \textbf{87.1} & \textbf{97.7}  & \textbf{94.6} & 71.7 & \textbf{87.8}  \\ \bottomrule
  \end{tabular}
  }
  \label{tab:1}
\end{table*}

\section{Experiments}

\noindent\textbf{Datasets.}
We evaluate our method on four popular few-shot action recognition datasets: HMDB51~\cite{kuehne2011hmdb}, UCF101~\cite{soomro2012ucf101}, Kinetics~\cite{carreira2017quo} and Something-Something V2 (SSv2)~\cite{goyal2017something}. HMDB51 contains 51 action categories, each containing at least 101 clips for a total of 6,766 video clips. For HMDB51, we adopt the same protocol as~\cite{zhang2020few} with 31/10/10 categories for train/val/test. UCF101 contains 101 action categories, each containing at least 100 clips for a total of 13,320 video clips. For UCF101, we also adopt the same protocol as~\cite{zhang2020few} with 70/10/21 categories for train/val/test. Kinetics contains 400 action categories with 400 or more clips for each category. For Kinetics, we adopt the same protocol as~\cite{zhu2018compound} with 64/12/24 categories for train/val/test. SSv2 contains 220,847 videos of fine-grained actions with only subtle differences between categories, and is regarded as a more challenging task. For SSv2, we adopt the same protocol as~\cite{cao2020few} with 64/12/24 categories for train/val/test.

\noindent\textbf{Baseline.}
Our work adopts STRM~\cite{thatipelli2022spatio} and TRX~\cite{perrett2021temporal} as baseline. STRM is a feature-based method based on TRX, and TRX applies CrossTransformer~\cite{doersch2020crosstransformers} to the few-shot action recognition scenario. CrossTransformer combines the information of support images and query images through an attention operation. TRX further samples ordered sub-sequences of video frames called tuples, and thus can capture higher-order temporal relationships. More details are demonstrated in the original article~\cite{perrett2021temporal}.

\noindent\textbf{Implementation details} 
For a better multimodal initialization, we use a pre-trained CLIP ResNet-50~\cite{he2016deep} as the visual backbone and a frozen CLIP text encoder based on a modified Transformer~\cite{vaswani2017attention} in~\cite{radford2019language}. Each video is re-scaled to height 256 and uniformly sampled $L=8$ frames as in~\cite{wang2016temporal}. We follow the TRX augmentation: random horizontal flipping and 224x224 crops in the meta-training stage, and only a center crop in the meta-test stage. We set $n_{temp}=16$, $d_{p}=d=1024$, $\Omega=\{2,3\}$ and multimodal enhanced hyper-parameter $\lambda=0.5$. According to ~\cite{rao2022denseclip}, we use AdamW~\cite{loshchilov2017decoupled} as our optimizer with a learning rate of $10^{-5}$ for HMDB51, UCF101, Kinetics and SSv2. We randomly sample 10000 training episodes for HMDB51, UCF101 and Kinetics, while 75000 training episodes for SSv2. We average gradients and backpropagate once every 16 iterations. In the meta-test stage, we employ the standard 5-way 5-shot evaluation on all four datasets. We randomly sample 10000 test episodes and report the average accuracy.

\begin{figure*}[t]
  \centering
  \subfloat[PRIDE gains by MORN.]{
    \includegraphics[width=0.37\linewidth]{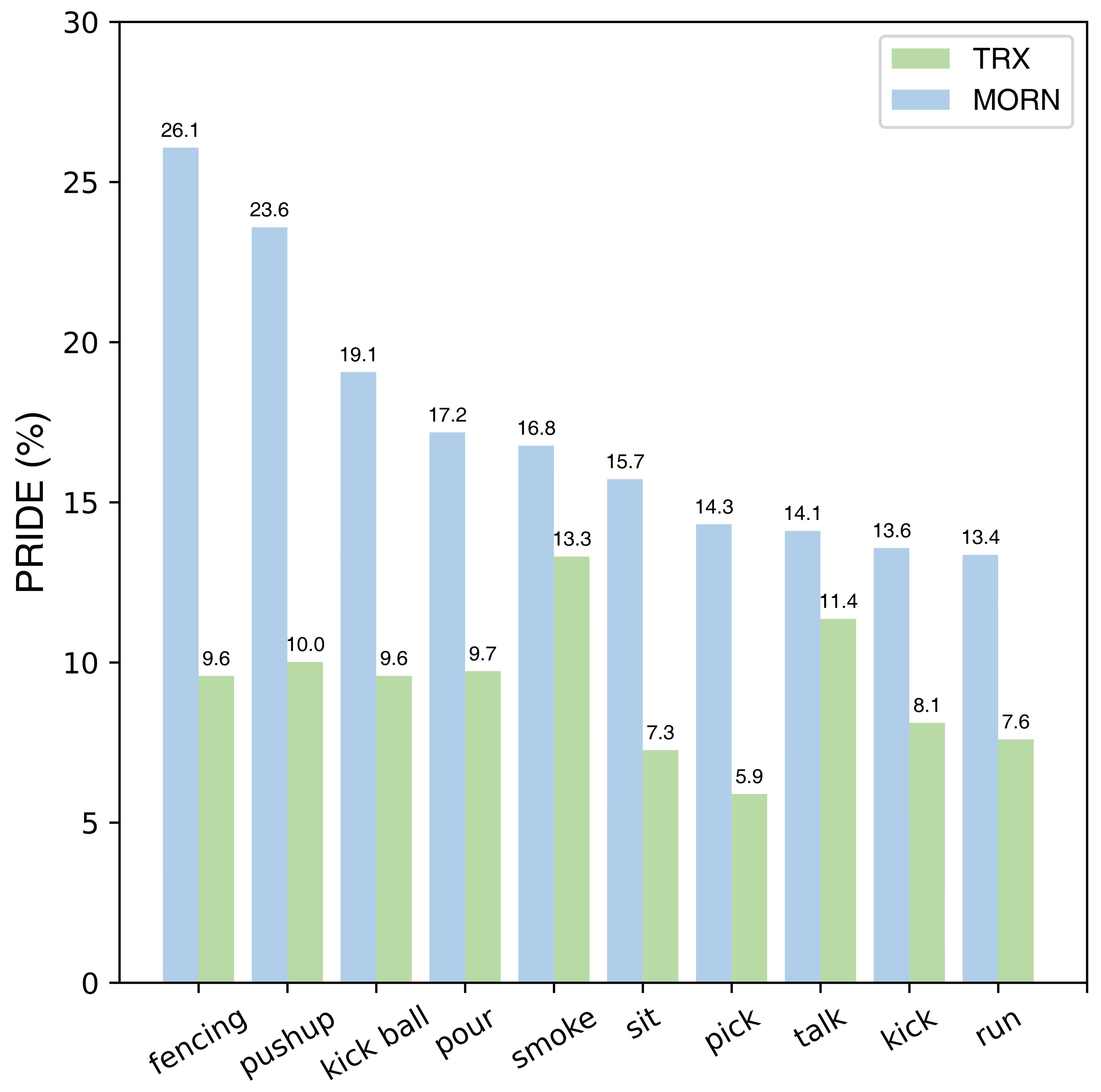}
    \label{fig3-a}
    }
  \subfloat[Accuracy gains by MORN.]{
    \includegraphics[width=0.38\linewidth]{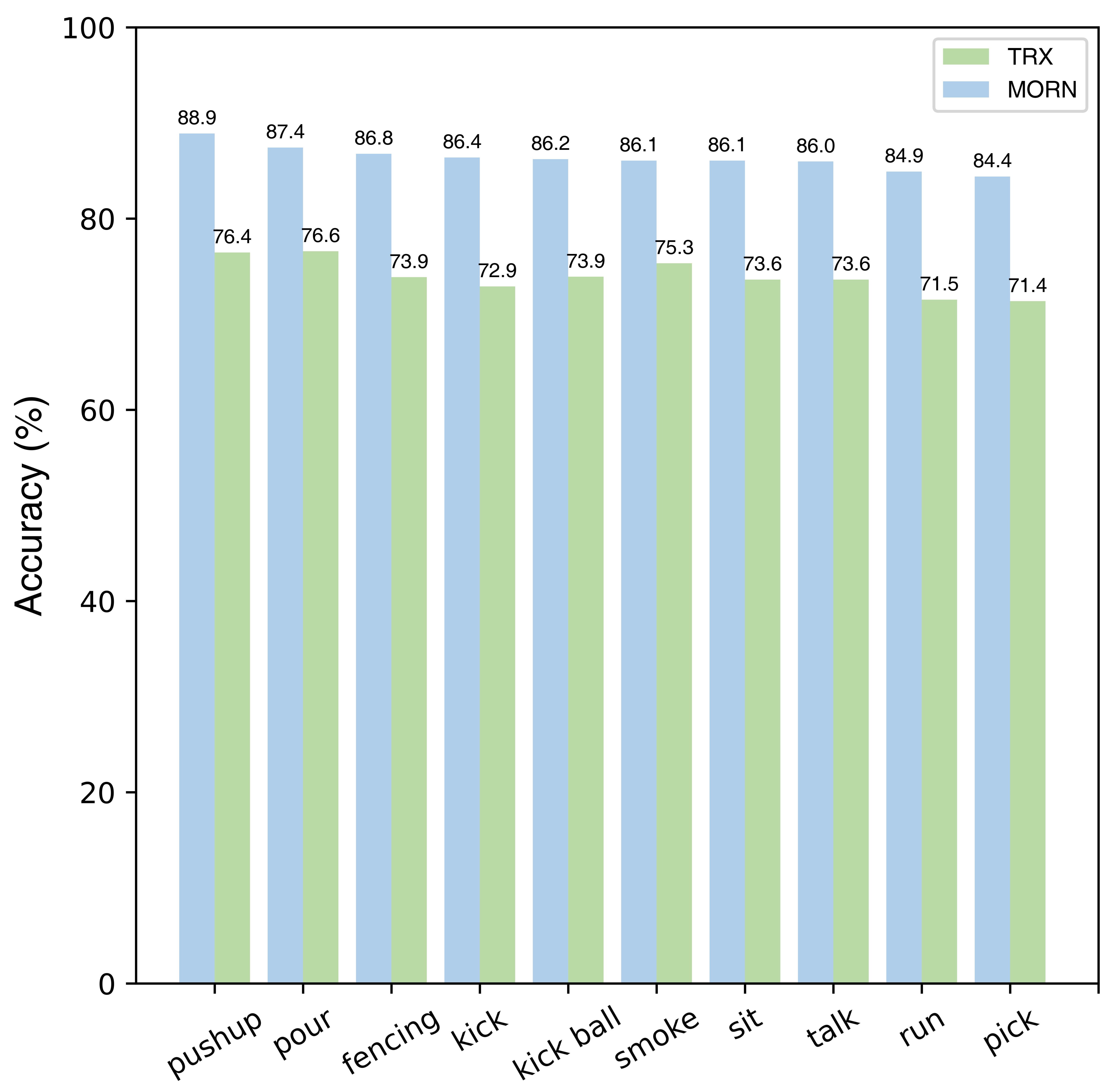}
    \label{fig3-b}
    }
  \caption{Performance gains of PRIDE and accuracy on HMDB51. MORN achieves PRIDE gains in (a) and accuracy gains in (b).} 
  \label{fig3}
\end{figure*}

\subsection{Comparison with State-of-the-arts}
\label{comparison with state-of-the-arts}

As shown in Table \ref{tab:1}, we apply MORN on STRM~\cite{thatipelli2022spatio} and TRX~\cite{perrett2021temporal}, and comprehensively compare four datasets for the standard 5-way 5-shot action recognition task with state-of-the-art methods. For fairness, we give preference to ResNet-50 or CLIP ResNet-50 in comparison if possible. Note that CLIP-FSAR~\cite{wang2023clip} and MA-CLIP~\cite{xing2023multimodal} are proposed later than us, while we still involve the two methods for a comprehensive comparison. CPMT~\cite{huang2022compound} uses both object features and a compound matching strategy, and achieves the best results on HMDB51 and SSv2 except ours. On UCF101 and Kinetics, both CLIP-FSAR~\cite{wang2023clip} and MA-CLIP~\cite{xing2023multimodal} use label texts as multimodal information to enhance prototypes, and achieve the best results respectively on the two datasets except ours. Among them, two methods (CLIP-FSAR and MA-CLIP) apply multimodal information and attempt to enhance the prototypes, which verify the effectiveness of multimodal information and prototypes preliminarily. Compared to previous SOTA, our MORN achieves the best results of 87.1\%, 97.7\%, 94.6\% and 73.6\% on all four datasets. Also, there is a significant improvement over STRM and TRX baseline with our MORN. When using weighted average as MPE module, STRM$^M$ achieves performance gains of 7.1\% on HMDB51, 3.8\% on Kinetics and 5.5\% on SSv2 over STRM, and TRX$^M$ also achieves similar performance gains. The improvement demonstrates the effectiveness of our multimodal prototype-enhanced strategy. 

To further improve the performance of MORN, we use a multi-head attention mechanism with 8 heads as MPE module. With a better MPE, the performance on HMDB51, UCF101 and Kinetics can be further improved, while the performance of STRM$^M$ on SSv2 declines on the contrary. It is probably because some label texts in SSv2 are so similar that it is hard to capture the useful semantic information, \emph{e.g.} \emph{pouring something into something} against \emph{pouring something onto something}. Plus, we notice the importance of multimodal initialization and introduce CLIP encoders. A more detailed ablation study of the MPE module and encoders will be illustrated in Section \ref{multimdoal encoders}. In summary, MORN achieves the best results of 87.1\% and 87.8\% with multi-head attention on average. For simplicity and fewer parameters, we apply TRX$^{M}$ with weighted average as our default setting of MORN.

\begin{table}[t]
  \centering
  \caption{Average PRIDE values across prototypes computed in $C_{novel}$ on HMDB51, UCF101, Kinetics and SSv2. *$^{C}$ denotes methods with CLIP visual encoder.}
  \begin{tabular}{@{}ccccc@{}}
  \toprule
  Method & HMDB51 & UCF101 & Kinetics & SSv2 \\\midrule
  STRM$^{C}$    & 10.0   & 17.5   & 16.5 & 8.6 \\
  Ours: STRM$^{M}$    & \textbf{17.1}   & \textbf{23.5}   & \textbf{26.5} & \textbf{48.2} \\\hline
  TRX$^{C}$    & 7.4   & 15.3   & 16.4 & 10.4 \\
  Ours: TRX$^{M}$   & \textbf{17.4}   & \textbf{24.4}   & \textbf{26.9} & \textbf{47.3} \\ \bottomrule
  \end{tabular}
  \label{tab:2}
\end{table}

\begin{figure}[t]
  \centering
  \subfloat[t-SNE of TRX on HMDB51.]{
    \includegraphics[width=0.47\linewidth]{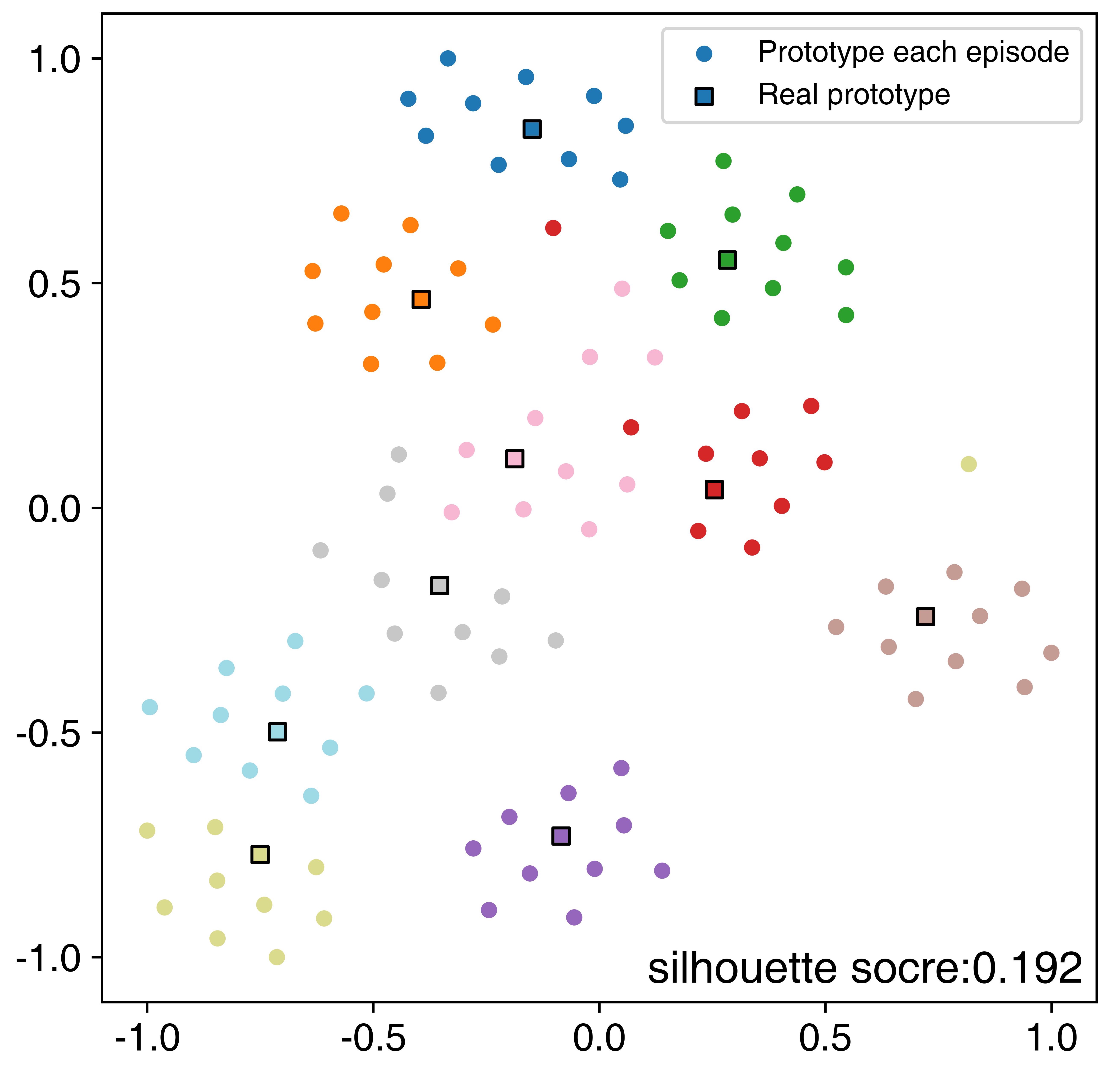}
    \label{fig4-a}
  }
  \hfill
  \subfloat[t-SNE of TRX$^{M}$ on HMDB51.]{
    \includegraphics[width=0.47\linewidth]{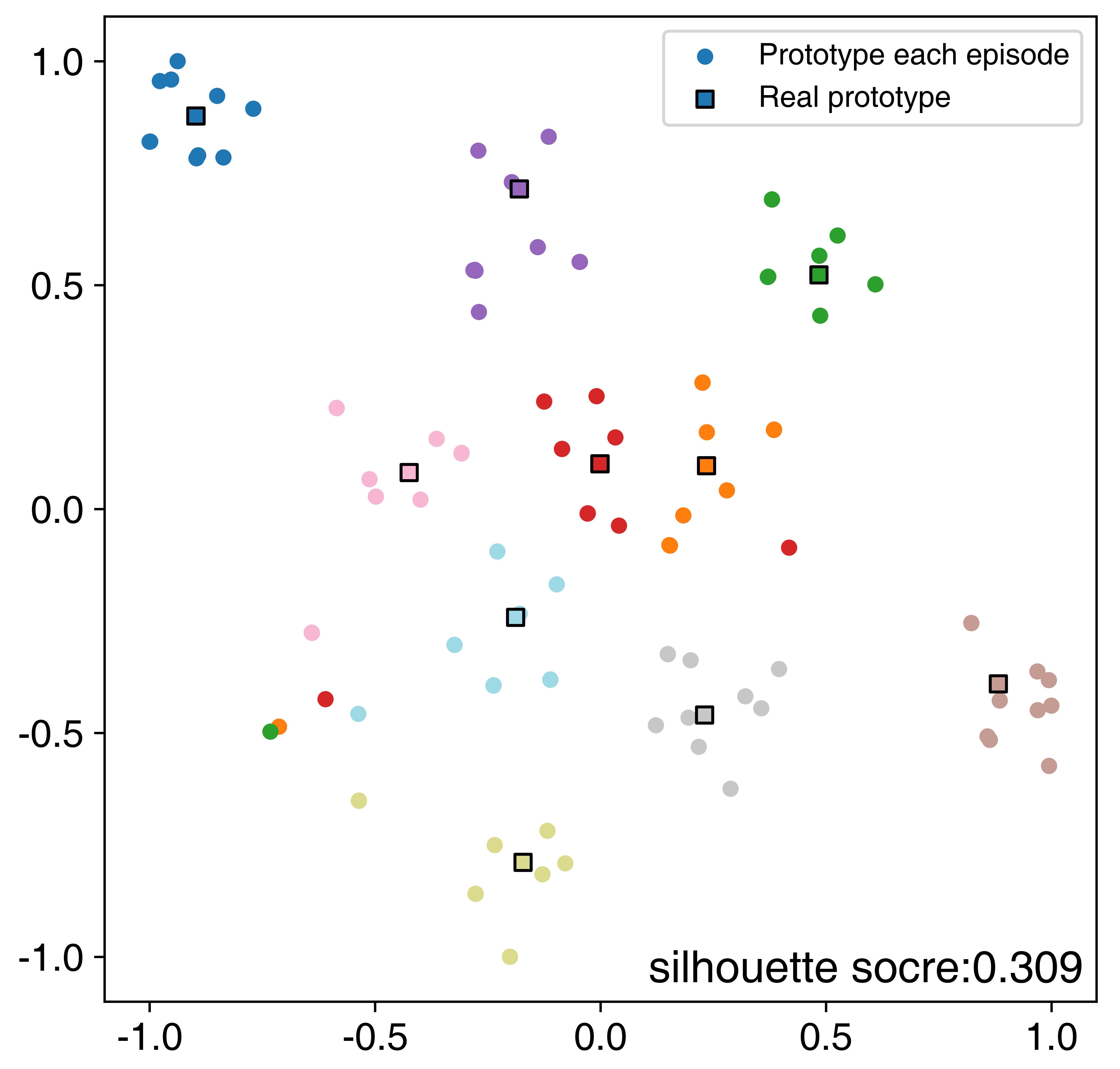}
    \label{fig4-b}
  }
  \caption{t-SNE~\cite{van2008visualizing} projection of prototypes of each episode and real prototypes on HMDB51.}
  \label{fig4}
\end{figure}

\begin{figure*}[t]
  \centering
  \includegraphics[width=0.7\linewidth]{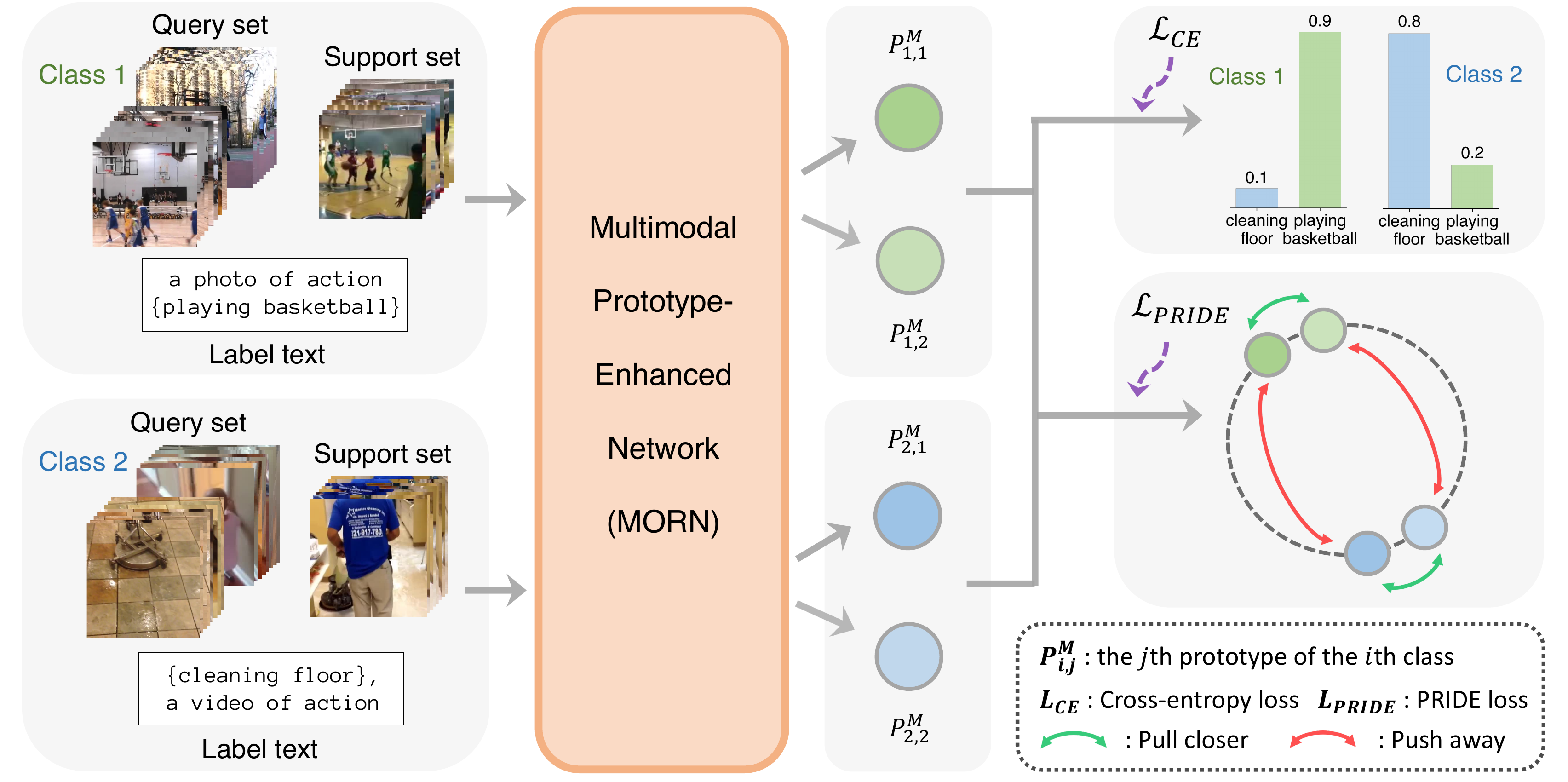}
  \caption{Overview of our proposed MORN with PRIDE loss on a 2-way 1-shot problem with 2 videos for each category in the query set. CE loss and PRIDE loss are combined through a learnable weight.}
  \label{fig5}
\end{figure*}

\subsection{Multimodal Prototype-Enhanced Analysis}

In this subsection, we explore the performance of our multimodal prototype-enhanced strategy. As mentioned earlier, we use PRIDE to evaluate the quality of prototypes. Firstly, we compute PRIDE and use it to verify the effectiveness of our MORN. Furthermore, we plug PRIDE into the training stage of MORN and achieve further performance gains.

Firstly, we randomly sample 10000 test episodes to compute accuracy and PRIDE. As shown in Figure \ref{fig3-a} and \ref{fig3-b}, MORN achieves significant gains on both PRIDE and accuracy on each category of HMDB51. Then, we conduct experiments of average PRIDE on all four datasets to verify the effectiveness of MORN. For a fair comparison, we replace the original ResNet-50 in TRX and STRM with a CLIP ResNet-50. As shown in Table \ref{tab:2}, there is significant improvement of STRM$^M$ and TRX$^M$ on PRIDE over STRM and TRX baseline. Specifically, STRM$^M$ achieves performance gains of 7.1\% on HMDB51, 6.0\% on UCF101, 10.0\% on Kinetics and 39.6\% on SSv2 over STRM, and TRX$^M$ also achieves similar PRIDE gains. Note that the improvement on SSv2 is particularly significant because the label texts in SSv2 are more complicated with more semantic information. According to the computation process, PRIDE can accurately reflect the category separability of prototypes. With MORN, we obtain higher PRIDE values over STRM and TRX. It turns out that MORN can improve the quality of prototypes, which verifies the effectiveness of MORN.

\begin{table}[t]
  \centering
  \caption{*$^{MP}$ denotes methods with our MORN and PRIDE loss. Plugging PRIDE into the training stage of MORN achieves further performance gains.}
  \begin{tabular}{@{}ccccc@{}}
  \toprule
  Method & HMDB51 & UCF101 & Kinetics & SSv2 \\\midrule
  STRM$^{M}$    & 84.4   & 94.6   & 90.5 & 73.6 \\
  STRM$^{MP}$    & \textbf{85.5}   & \textbf{95.6}   & \textbf{91.2} & \textbf{73.9} \\\hline
  TRX$^{M}$    & 86.3   & \textbf{96.9}   & 91.6 & 71.1 \\
  TRX$^{MP}$   & \textbf{86.4}   & \textbf{96.9}   & \textbf{91.9} & \textbf{72.6} \\ \bottomrule
  \end{tabular}
  \label{tab:3}
\end{table}

Intuitively, a method with a higher PRIDE value means a better discriminating performance against various categories. We compute real prototypes and randomly sample 10 prototypes in each category. Then, we visualize the clusters by t-SNE~\cite{van2008visualizing} on HMDB51 as shown in Figure \ref{fig4}. TRX$^{M}$ is our default MORN. Compared to TRX, MORN has a better category separability, and prototypes are closer to the real prototypes. To quantify the results, we compute the silhouette values of the two methods. MORN outperforms TRX by 0.117 on HMDB51, showing that MORN computes more representative prototypes for classification.

To further explore the function of PRIDE, we plug it into the training stage of MORN as a loss as shown in Figure \ref{fig5}. Formally, PRIDE loss is implemented through InfoNCE~\cite{oord2018representation} with prototypes:
${\mathcal L}_{PRIDE}=-\log\frac{\exp(P_{i}(x)\!\cdot\!P^{real}_{i}/\tau)}{\sum_{j=1}^{N_{novel}}\exp(P_{i}(x)\!\cdot\!P^{real}_{j}/\tau)}$.

We aim at increasing the similarity of prototypes in the same category and decreasing that of prototypes in different categories. CE loss and PRIDE loss are combined through a learnable weight. Since PRIDE indicates the quality of prototypes, it is supposed to assist the training process. In Table \ref{tab:3}, STRM$^{MP}$ and TRX$^{MP}$ achieve extra performance gains over STRM$^{M}$ and TRX$^{M}$. The results verify both the rationality of PRIDE and the effectiveness of our MORN. 

\subsection{Ablation Study}
\label{Ablation Study}

\begin{table}[t]
  \centering
  \caption{Ablation study of various encoders on HMDB51, UCF101, Kinetics and SSv2. *$^{RB}$ denotes methods with ResNet-50 and BERT-large.}
  \resizebox{\linewidth}{!}{
  \begin{tabular}{@{}cccccc@{}}
  \toprule
  Method & Backbone & HMDB51 & UCF101 & Kinetics & SSv2 \\\midrule
  STRM   & \multirow{2}{*}{ResNet-50} & 77.3 & \textbf{96.9} & 86.7 & 68.1 \\
  TRX    &  & 75.6 & 96.1 & 85.9 & 64.6 \\\hline
  STRM$^C$ & \multirow{2}{*}{CLIP ResNet-50} & 79.3 & 94.7 & 86.0 & 62.9 \\
  TRX$^C$   &  & 77.9   & 95.9    & 86.7      & 64.4 \\\hline
  STRM$^{RB}$  & \multirow{2}{*}{ResNet-50 + BERT} & 73.4   & 94.8   & 83.2    & 59.3 \\
  TRX$^{RB}$  &  & 72.0   & 95.2    & 83.4    & 58.8 \\\hline
  STRM$^{M}$    & \multirow{2}{*}{\begin{tabular}{@{}c@{}} CLIP ResNet-50 + \\CLIP text encoder \end{tabular}} & 84.4   & 94.6   & 90.5 & \textbf{73.6} \\
  TRX$^{M}$    &  & \textbf{86.3} & \textbf{96.9} & \textbf{91.6} & 71.1 \\\bottomrule
  \end{tabular}
  }
  \label{tab:4}
\end{table}

\noindent\textbf{Multimodal encoders.}
\label{multimdoal encoders}
We employ label texts as an extra modality, and an intuitive idea is to use BERT~\cite{devlin2019bert} as the text encoder. However, the performance of STRM$^{RB}$ and TRX$^{RB}$ declines a lot as shown in Table \ref{tab:4}. We refer it to an inadequate multimodal initialization. To verify this, we introduce CLIP encoders with a good multimodal initialization and our prototype-enhanced strategy. STRM$^{M}$ and TRX$^{M}$ achieve best results and significantly outperform STRM$^{RB}$ and TRX$^{RB}$. Furthermore, to verify the importance of the text flow, we only replace the original ResNet-50 with a CLIP ResNet-50 and obtain STRM$^{C}$ and TRX$^{C}$. The performance varies on four datasets, and is worse than that of STRM$^{M}$ and TRX$^{M}$, demonstrating the importance of multimodal information. In addition, we show the advantage of prototype-enhanced strategy over feature-enhanced strategy in Figure \ref{fig1}. Combining the two aspects, the necessity and effectiveness of our MORN are verified.

\begin{figure}[t]
  \centering
  \includegraphics[width=0.65\linewidth]{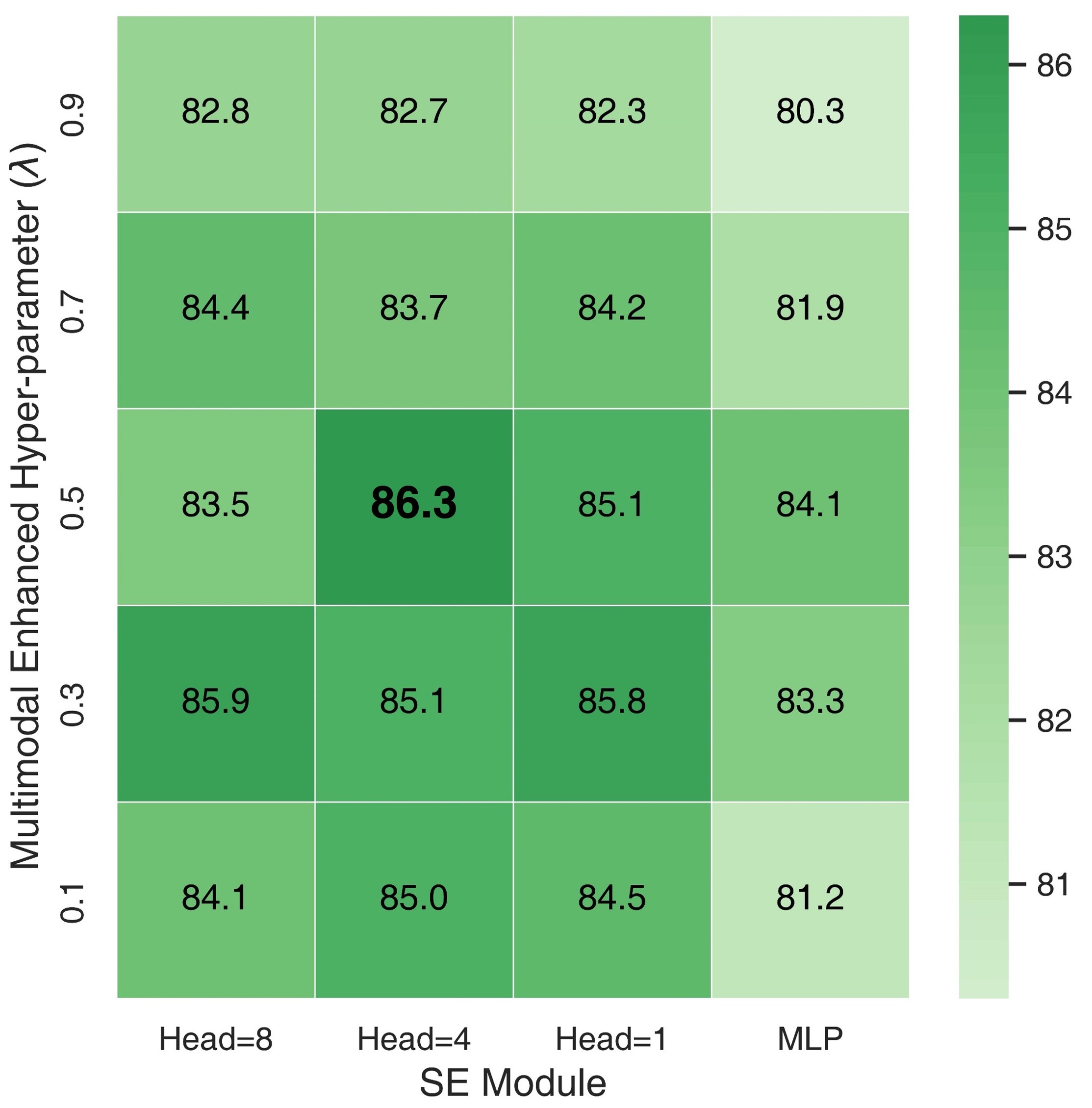}
  \caption{Ablation study of varying the SE module and multimodal enhanced hyper-parameter ($\lambda$) on HMDB51.}
  \label{fig6}
\end{figure}

\begin{table}[t]
  \centering
  \caption{Ablation study of the MPE module on HMDB51, UCF101, Kinetics and SSv2. ``Concat'' indicates a concatenation operation and ``Head'' indicates the head number in multi-head attention mechanism. The bold font and underline indicate the best and second-best results respectively.}
  \begin{tabular}{@{}ccccc@{}}
  \toprule
  MPE module & HMDB51 & UCF101 & Kinetics & SSv2 \\\midrule
  Concat + MLP    & 62.0   & 79.9   & 76.6 & 58.3 \\
  MLP + Concat    & 72.4   & 86.3   & 80.5 & 60.6 \\
  Weighted average    & \underline{86.3}   & \underline{96.9}   & 91.6 & 71.1 \\
  Head = 1    & 81.0   & 95.5   & 91.8 & \textbf{72.6} \\
  Head = 4   & 85.1   & \underline{96.9}   & \underline{94.1} & 70.3 \\ 
  Head = 8   & \textbf{87.1}   & \textbf{97.7}   & \textbf{94.6} & \underline{71.7} \\
  \bottomrule
  \end{tabular}
  \label{tab:5}
\end{table}

\noindent\textbf{Design choices.} As shown in Figure \ref{fig6}, we compare various choices of the SE module and multimodal enhanced hyper-parameter ($\lambda$) on HMDB51. When using a two-layer MLP with a 1024-dimensional hidden layer, we obtain a low accuracy. It is mainly because MLP has a simple architecture and thus fails to fully explore the semantic information of label texts. Therefore, we focus on the multi-head attention mechanism~\cite{vaswani2017attention}. As the head number increases, the fitting ability of the SE module improves as well. However, the accuracy fluctuates with different head numbers and $\lambda$ choices. 
We need to balance the trade-off between head numbers and $\lambda$ choices. According to our results, we use a multi-head attention mechanism with 4 heads as the SE module and set $\lambda$ as 0.5 in our further experiments. 

For the MPE module, applying a multi-head attention mechanism can further improve the performance of MORN as shown in Table \ref{tab:5}. It turns out that the multimodal information can further enhance the prototypes with a better multimodal combination strategy. MORN achieves the best results with 8 heads on HMDB51, UCF101 and Kinetics, while with 1 head on SSv2. Although weighted average is relatively simple with no parameters for training, our MORN with weighted average can also achieve the second-best results of 86.3\% on HMDB51 and 96.9\% on UCF101, which has already outperformed those of prior methods. For simplicity and fewer parameters for training, we use weighted average as our default MPE module. 

\begin{table}[t]
  \centering
  \caption{Ablation study of the SE module and CLIP text encoder on HMDB51, UCF101, Kinetics and SSv2.}
  \resizebox{\linewidth}{!}{
  \begin{tabular}{@{}cccccc@{}}
  \toprule
  \begin{tabular}{@{}c@{}} SE module\\(exist or not) \end{tabular} & \begin{tabular}{@{}c@{}} CLIP text encoder\\(freeze or not) \end{tabular} & HMDB51 & UCF101 & Kinetics & SSv2 \\\midrule
  \ding{56}   & \ding{56}   & 84.5   & 96.1   & 88.8 & 72.5\\
  \ding{56}   & \ding{52}   & 83.9   & 96.4   & 90.6 & \textbf{72.8}\\ 
  \ding{52}   & \ding{56}   & 76.7   & 93.2   & 87.2 & 62.1\\
  \ding{52}   & \ding{52}   & \textbf{86.3}   & \textbf{96.9}   & \textbf{91.6} & 71.1\\
  \bottomrule
  \end{tabular}
  }
  \label{tab:6}
\end{table}

\noindent\textbf{SE module and CLIP text encoder.} 
In the text flow, we first introduce a frozen CLIP text encoder. Then, a semantic-enhanced (SE) module is used to enhance the semantic information. Firstly on SSv2, MORN with a frozen CLIP text encoder is enough to achieve the best result of 72.8\% as shown in Table \ref{tab:6}. When coming to HMDB51, UCF101 and Kinetics, an SE module further achieves performance gains, and MORN with the SE module and a frozen CLIP text encoder achieves the best results. On the one hand, the function of the SE module is similar to that of the enrichment module in ~\cite{thatipelli2022spatio} to obtain reliable semantic information. On the other hand, a frozen language encoder is proven to be effective in several works~\cite{tsimpoukelli2021multimodal, rao2022denseclip} and our results further prove it. When we remove the SE module, the results are lower than MORN whether we freeze the CLIP text encoder or not. Besides, using the SE module and fine-tuning the CLIP text encoder leads to the lowest results on all three datasets. It is mainly because the overfitting problem becomes more serious as parameters increase. When coming to SSv2, MORN with a frozen CLIP text encoder but without the SE module achieves the best result. In general, MORN achieves significant performance gains with our design.

\section{Conclusion}

We propose a novel Multimodal Prototype-Enhanced Network (MORN) for few-shot action recognition. Our MORN uses the semantic information of label texts as multimodal information to compute more representative prototypes, and achieves state-of-the-art results. We also define Prototype Similarity Difference (PRIDE) to evaluate the quality of prototypes. By conducting experiments on PRIDE, we verify our improvement on prototypes. Besides, plugging PRIDE into the training stage of MORN can achieve further performance gains, verifying the rationality of PRIDE and the importance of prototypes. In summary, we'd like to shed light on a view: \emph{multimodal prototypes with high quality are extremely important in few-shot action recognition}.

\section*{Acknowledgements}
This work was partly supported by the National Natural Science Foundation of China (Grant No. U1903213) and the Shenzhen Science and Technology Program (JSGG20220831093004008).

\bibliographystyle{ACM-Reference-Format}
\balance
\bibliography{sample-base}

\end{document}